\documentclass[conference]{IEEEtran}
\IEEEoverridecommandlockouts
\usepackage{cite}
\usepackage{amsmath,amssymb,amsfonts}
\usepackage{algorithmic}
\usepackage{graphicx}
\usepackage{textcomp}
\usepackage[table]{xcolor}

\usepackage{subcaption}
\usepackage{marvosym}
\usepackage{multirow}
\usepackage{booktabs}
\usepackage{comment}

\makeatletter
\newcommand{\linebreakand}{%
  \end{@IEEEauthorhalign}
  \hfill\mbox{}\par
  \mbox{}\hfill\begin{@IEEEauthorhalign}
}
\makeatother

\def\BibTeX{{\rm B\kern-.05em{\sc i\kern-.025em b}\kern-.08em
    T\kern-.1667em\lower.7ex\hbox{E}\kern-.125emX}}
\begin{document}

\title{EPContrast: Effective Point-level Contrastive Learning for Large-scale Point Cloud Understanding}

\author{\IEEEauthorblockN{1\textsuperscript{st} Zhiyi Pan}
\IEEEauthorblockA{\textit{School of Electronic and Computer Engineering,} \\ 
\textit{Peking University} \\
\textit{\& Peng Cheng Laboratory}\\
Shenzhen, China \\
panzhiyi@stu.pku.edu.cn}
\and
\IEEEauthorblockN{2\textsuperscript{nd} Guoqing Liu}
\IEEEauthorblockA{\textit{\quad\qquad\qquad\qquad Minieye Inc.\qquad\qquad\qquad\quad} \\
Shenzhen, China \\
guoqing@minieye.cc}
\linebreakand
\IEEEauthorblockN{3\textsuperscript{rd} Wei Gao}
\IEEEauthorblockA{\textit{School of Electronic and Computer Engineering,} \\ \textit{Peking University} \\
Shenzhen, China \\
gaowei262@pku.edu.cn}
\and
\IEEEauthorblockN{4\textsuperscript{th} Thomas H. Li \textsuperscript{\rm\Letter}\thanks{\Letter\ Thomas H. Li is the corresponding author. This work was supported in part by Natural Science Foundation of China (No. 62172021), in part by Shenzhen Science and Technology Program (KQTD20180411143338837), and in part by National Key R\&D Program of China (No.2020AAA0103501).}}
\IEEEauthorblockA{\textit{School of Electronic and Computer Engineering,} \\ \textit{Peking University} \\
Shenzhen, China \\
thomas@pku.edu.cn}
}

\maketitle

\begin{abstract}
The acquisition of inductive bias through point-level contrastive learning holds paramount significance in point cloud pre-training. However, the square growth in computational requirements with the scale of the point cloud poses a substantial impediment to the practical deployment and execution. To address this challenge, this paper proposes an \textbf{E}ffective \textbf{P}oint-level \textbf{Contrast}ive Learning method for large-scale point cloud understanding dubbed \textbf{EPContrast}, which consists of AGContrast and ChannelContrast. In practice, AGContrast constructs positive and negative pairs based on asymmetric granularity embedding, while ChannelContrast imposes contrastive supervision between channel feature maps. EPContrast offers point-level contrastive loss while concurrently mitigating the computational resource burden. The efficacy of EPContrast is substantiated through comprehensive validation on S3DIS and ScanNetV2, encompassing tasks such as semantic segmentation, instance segmentation, and object detection. In addition, rich ablation experiments demonstrate remarkable bias induction capabilities under label-efficient and one-epoch training settings.
\end{abstract}
\section{Introduction}
In recent years, the scale of point cloud scenes for development and application has exhibited a notable upsurge~\cite{armeni20163d,dai2017scannet,behley2019semantickitti}. The comprehension of large-scale point clouds increasingly depends on pre-training. By training with an unsupervised pretext task, the network learns the inherent inductive bias from massive unlabeled point clouds, which significantly improves the performance of downstream tasks after fine-tuning.

\begin{figure}[htbp]
	\centering
	\begin{subfigure}{\linewidth}
		\centering
		\includegraphics[width=\linewidth]{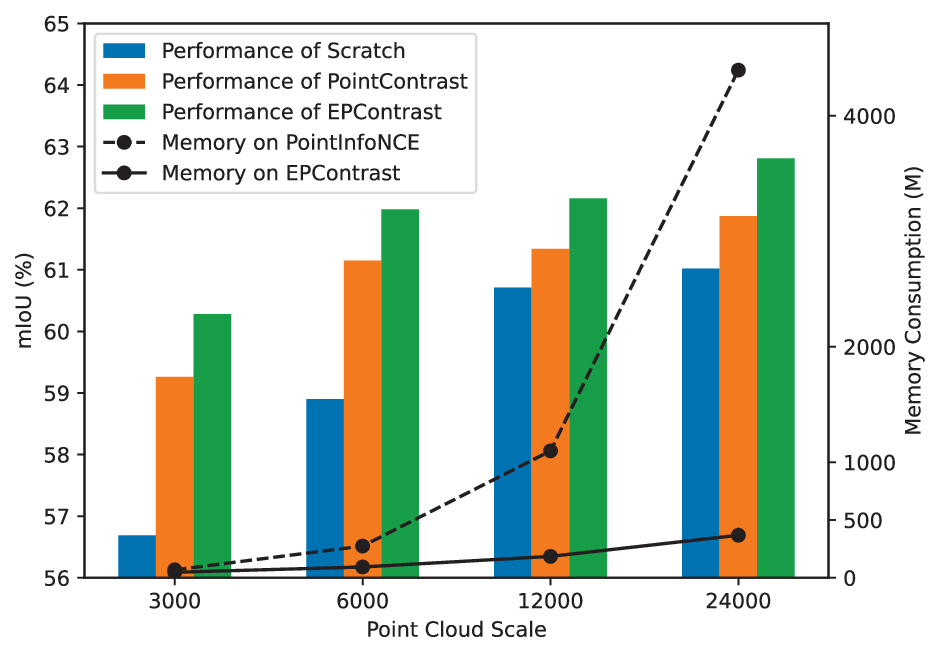}
	\end{subfigure}

	\centering
	\begin{subfigure}{\linewidth}
		\centering
		\includegraphics[width=\linewidth]{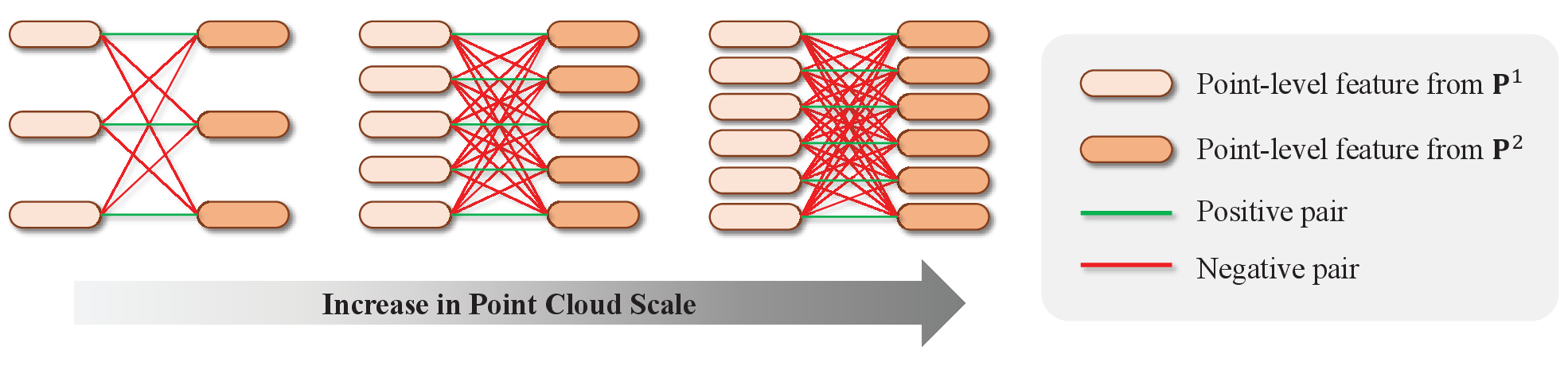}
	\end{subfigure}
 
	\caption{(\textbf{Top}) The semantic segmentation performance on Scratch, PointContrast, and our proposed EPContrast, and memory consumption on PointInfoNCE (used in PointContrast) and EPContrast. (\textbf{Bottom}) Visualization of point-level contrastive learning with increasing point cloud scale.}
	\label{fig:main}
\end{figure}

Contrastive learning, as a prominent pre-training paradigm~\cite{xie2020pointcontrast}, endows the model with inductive bias through the construction of positive and negative pairs for point cloud samples. However, as shown in Fig.~\ref{fig:main} the number of negative pairs generated via point-level contrastive learning escalates quadratically with the augmentation of the scene scale. Consequently, a substantial allocation of computational resources is requisite, thereby rendering the direct application of contrastive learning to large-scale point cloud scene understanding a formidable challenge.

To tackle this challenge, various strategies have been employed within existing contrastive learning methods such as randomly sampling~\cite{xie2020pointcontrast, zhang2021self,chen20224dcontrast}, segment-level contrastive learning~\cite{nunes2022segcontrast,hou2021exploring} and scene-level contrastive learning~\cite{sanghi2020info3d, afham2022crosspoint, li2023tothepoint}. Nevertheless, these strategies compromise on the loss of crucial information, the relaxation of granularity, and the introduction of additional computations, all of which lead to suboptimal pre-training performance.

In this paper, we introduce an \textbf{E}ffective \textbf{P}oint-level \textbf{Contrast}ive learning method (EPContrast) for large-scale point cloud understanding. On the one hand, we present AGContrast, which considers asymmetric granularity embedding in contrastive learning pairs. AGContrast seeks the corresponding segment features for each point within a point cloud scene, thereby preserving point-level contrastive loss and delivering fine-grained supervision for subsequent point cloud understanding network initialization with a linear source utilization. On the other hand, we introduce ChannelContrast to efficiently utilize computational resources and address potential channel redundancy concerns~\cite{li2023scconv}. ChannelContrast activates each dimension of network embedding, treating feature maps within the same channel as positive pairs and those in different channels as negative pairs. By ensuring a certain degree of orthogonality between feature maps, ChannelContrast mitigates high-dimensional feature redundancy during pre-training. As illustrated in Fig.~\ref{fig:main}, EPContrast operates at a fine-grained level as all points within the scene contribute to the computation of the contrastive loss.

To assess the effectiveness of EPContrast in pre-training, we conduct comprehensive experiments on various point cloud understanding tasks using different backbone networks and large-scale datasets. Remarkably, EPContrast consistently outperforms other contrastive learning approaches across all configurations. This underscores the superior pre-training efficacy of EPContrast in enhancing the performance of point cloud understanding models. The ablation study provides comprehensive insights into the benefits of AGContrast and ChannelContrast. Moreover, we evaluate the performance of our approach under one-epoch training scenarios~\cite{tong2023emp} and label-efficient experimental setups~\cite{hou2021exploring}, showcasing the potential of EPContrast for both few-shot learning and weakly supervised learning.

The contribution of this paper can be summarized in the following four aspects:
\begin{itemize}
    \item We expose the deployment limitations of contrastive learning in large point cloud scenarios and propose a novel low memory consumption and performance-efficient contrastive learning method, EPContrast.
    \item We propose asymmetric granularity contrast learning (AGContrast), which builds supervised signals for contrast learning between point-level and segment-level features.
    \item The ChannelContrast is introduced to increase the representative ability between feature maps, by ensuring a certain degree of orthogonality.
    \item Through validation on various downstream tasks, EPContrast demonstrates significant performance benefits in most scenarios.
\end{itemize}

\section{Related Work}
The remarkable success of contrastive learning in 2D vision inspires the pre-training research for point clouds. Contrastive learning in 3D vision treats a scene, a segment, or a point as an instance. To realize the unsupervised discriminative pretext task learning, it brings the query instance closer to the positive key instances, while pushing it farther away from the negative key instances in the latent space. According to the granularity of contrastive learning, it can be categorized into scene-level, segment-level, and point-level contrastive learning.

\noindent \textbf{Scene-level contrastive learning.} Info3d~\cite{sanghi2020info3d} maximizes the mutual information between the object and its localization in contrastive learning, based on the hypothesis that objects and localities are semantically consistent. Crosspoint~\cite{afham2022crosspoint} constructs cross-modal contrastive learning between point clouds and images by projecting 3D point clouds onto 2D planes. ToThePoint~\cite{li2023tothepoint} maximizes the agreement between the scene-level global features and features discarded by max-pooling. Due to the coarse granularity of contrastive supervision, scene-level contrastive learning is hard to obtain discriminative inductive bias from the whole point cloud scene when confronted with large-scale point cloud scenes, which is only applicable to low-level downstream tasks (\emph{e.g.}, object classification, reconstruction) in simple point cloud scenarios.

\noindent \textbf{Point-level contrastive learning.} To implement contrast learning for high-level downstream tasks (\emph{e.g.}, semantic segmentation, object detection), several works propose point-level contrastive learning that enables pre-trained networks to learn inductive bias from the embedding distribution of points. For instance, MID~\cite{wang2021unsupervised} utilizes both scene-level and point-level contrastive loss for shape analysis. PointContrast~\cite{xie2020pointcontrast} applies point-level contrastive learning to the point cloud scene at two different viewpoints, while DepthContrast~\cite{zhang2021self} imposes point-level contrastive loss between the point and voxel representations of the 3D scene. Point-level contrastive learning sample negative pairs to avoid unaffordable memory consumption, but this also hinders the training efficiency.

\noindent \textbf{Segment-level contrastive learning.} Segment-level contrastive learning attempts to trade off fine-grained supervision against low memory consumption. For example,  SegContrast~\cite{nunes2022segcontrast} alleviates this dilemma by utilizing segment embedding as queries and keys. Recently, Contrastive Scene Contexts~\cite{hou2021exploring} leverages the ShapeContext local descriptor to partition the space and applies the contrastive learning to spatial chunks individually. Despite the significant reduction in resource usage, the effectiveness of contrastive learning is negatively impacted by reducing the granularity of samples.

Distinct from existing methods, we construct asymmetric granularity for contrastive learning by considering points and segments on the latent space as queries and keys, respectively. In addition, we introduce channel contrastive learning to alleviate channel redundancy in large-scale point cloud scenarios.

\begin{figure*}[htbp]
    \centering
    \includegraphics[width=\linewidth]{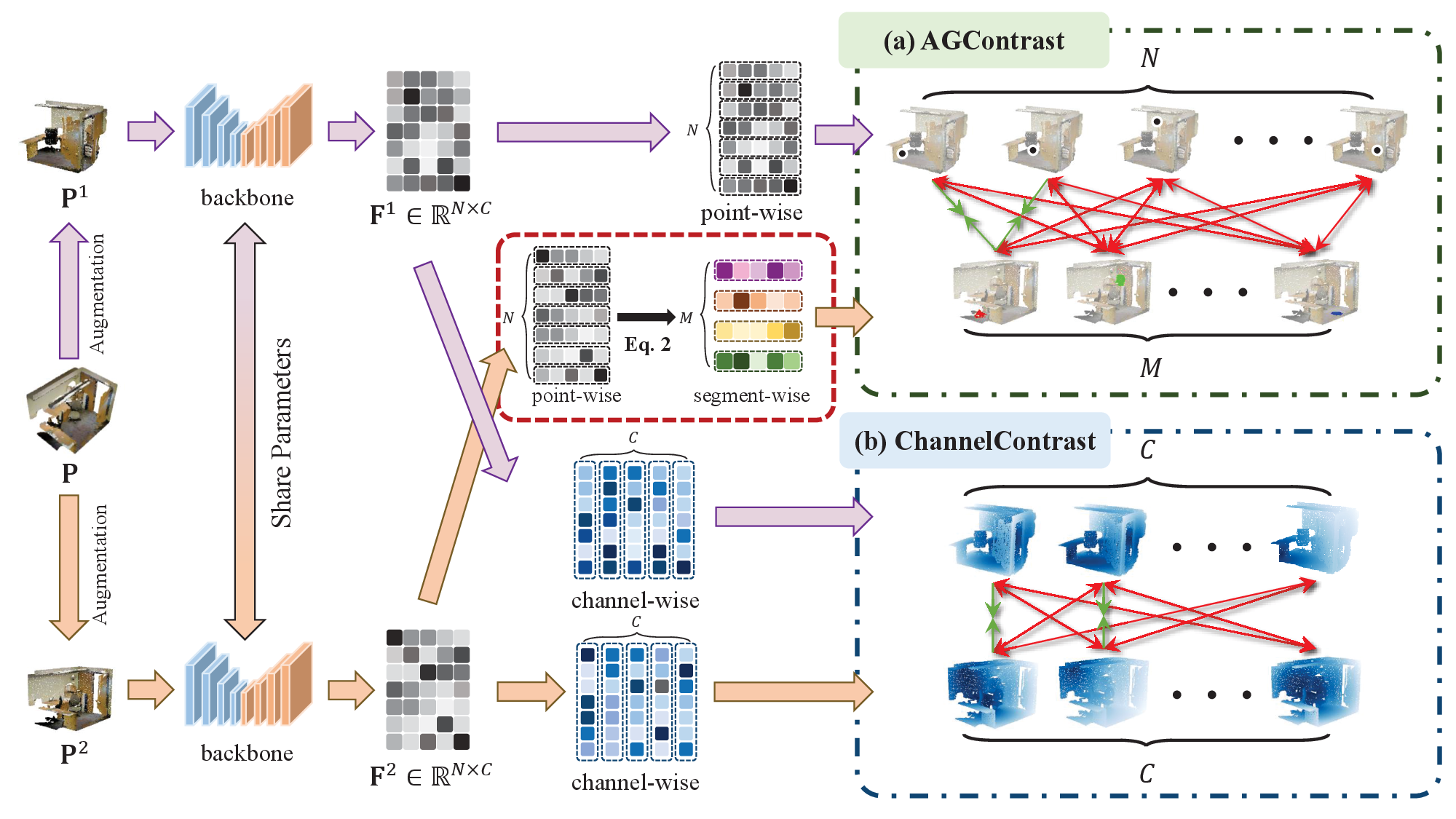}
    \caption{The pre-training framework with EPContrast, which consists of (a) AGContrast and (b) ChannelContrast.}
    \label{fig:network}
\end{figure*}

\section{Method}
In this section, we first overview the mathematical description of point-level contrastive learning. Subsequently, we introduce our method EPContrast, including the definitions of AGContrast and ChannelContrast. We conclude this section by presenting the overall loss function.

\subsection{Overview}
Assuming a point cloud scene in the dataset is $\mathbf{P}\in \mathbb{R}^{N\times D}$, where $N$ denotes the number of points in the scene, and $D$ denotes the dimension of features attached to the points. To facilitate point-level contrastive learning, it is essential to generate distinct representations for the point cloud $\mathbf{P}$ to establish correspondences. A more versatile approach to creating a pair of point cloud representations is data augmentation. We denote the augmented point cloud pairs as $\mathbf{P}^1$ and $\mathbf{P}^2$. The corresponding network embeddings are $\mathbf{F}^1$ and $\mathbf{F}^2\in \mathbb{R}^{N\times C}$, where $C$ denotes the dimension of the embedding. For PointInfoNCE in PointContrast~\cite{xie2020pointcontrast}, the set of positive pairs $\mathcal{P}=\{(i,i)\ |\ i=1,2,\cdots,N\}$ between $\mathbf{P}^1$ and $\mathbf{P}^2$, and the set of negative pairs $\mathcal{N}=\{(i,j)\ |\ i\neq j\}$. The point-level contrastive loss can be defined as
\begin{equation}
    \mathcal{L}_{\rm{PC}}=-\sum_{(i,i)\in\mathcal{P}}\log \frac{\exp (\mathbf{f}^1_i\cdot {\mathbf{f}^2_i}^\top/\tau)}{\sum_{(i,j)\in \mathcal{N}}\exp (\mathbf{f}^1_i\cdot {\mathbf{f}^2_j}^\top/\tau)},
\end{equation}
where $\mathbf{f}_i\in \mathbb{R}^{1\times C}$ denotes the feature of the $i$-th point in $\mathbf{F}$, and $\tau$ denotes the temperature that controls the smoothness of the softmax distribution. 

Since the size of the negative pair set $\mathcal{N}$ is $N^2-N$, the space complexity of point-level contrastive learning is $O(N^2)$. When confronted with large-scale point clouds, it is often necessary to randomly sample the points, which simultaneously loses some critical negative pairs supervision information. Therefore, we propose an \textbf{E}ffective \textbf{P}oint-level \textbf{Contrast}ive learning (EPContrast) for large-scale point cloud understanding, which is illustrated in Fig.~\ref{fig:network}.

\begin{table*}[htbp]
\caption{Detailed downstream task setups for contrastive learning in our implementation.}
\label{tab:settings}
\resizebox{\linewidth}{!}{
\begin{tabular}{c|c|c|ccccccc|c}

\toprule[3\arrayrulewidth]
Task                                                                             & Dataset                  & Backbone                                                                    & Phase     & Batch & Training Epoch & LR    & LR Decay & Optimizer & GPU(s) & Improve                                                                   \\ \hline
\multirow{2}{*}{\begin{tabular}[c]{@{}c@{}}Semantic\\ Segmentation\end{tabular}} & \multirow{2}{*}{S3DIS}   & \multirow{2}{*}{PointNet++}                                                 & pre-train & 8     & 100 epochs     & 0.01  & cosine   & AdamW     & 1 V100 & \multirow{2}{*}{\begin{tabular}[c]{@{}c@{}}+1.8\%\\ mIoU\end{tabular}}    \\ \cline{4-10}
                                                                                 &                          &                                                                             & fine-tune & 8     & 50 epochs      & 0.01  & cosine   & AdamW     & 1 V100 &                                                                           \\ \hline
\multirow{2}{*}{\begin{tabular}[c]{@{}c@{}}Object\\ Detection\end{tabular}}      & \multirow{2}{*}{ScanNet} & \multirow{2}{*}{VoteNet}                                                    & pre-train & 32    & 30K iters      & 0.1   & step     & SGD       & 4 V100 & \multirow{2}{*}{\begin{tabular}[c]{@{}c@{}}+4.4\%\\ mAP@0.5\end{tabular}} \\ \cline{4-10}
                                                                                 &                          &                                                                             & fine-tune & 32    & 180 epochs     & 0.001 & step     & SGD       & 1 V100 &                                                                           \\ \hline
\multirow{2}{*}{\begin{tabular}[c]{@{}c@{}}Instance\\ Segmentation\end{tabular}} & \multirow{2}{*}{ScanNet} & \multirow{2}{*}{\begin{tabular}[c]{@{}c@{}}Sparse\\ Res-U-Net\end{tabular}} & pre-train & 32    & 30K iters      & 0.1   & step     & SGD       & 4 V100 & \multirow{2}{*}{\begin{tabular}[c]{@{}c@{}}+1.7\%\\ mAP@0.5\end{tabular}} \\ \cline{4-10}
                                                                                 &                          &                                                                             & fine-tune & 48    & 10K iters      & 0.1   & poly.    & SGD       & 8 V100 &                   \\
\toprule[3\arrayrulewidth]                                                                                 
\end{tabular}
}
\end{table*}

\subsection{AGContrast}
To alleviate the space complexity associated with point-level contrastive learning, we introduce an asymmetric granularity contrastive loss involving both points and segments, as shown in Fig.~\ref{fig:network} (a). The asymmetric form of contrastive supervision encourages the network to learn inductive biases between different levels of features, resulting in enhanced generalization and robustness in large-scale point cloud scene understanding. To this end, we first perform unsupervised superpoint segmentation based on spatial and color information to obtain the segment set $\{\mathbf{S}_\alpha\ |\ \alpha=1,2,\cdots, M\}$, which satisfies $\cap_\alpha \mathbf{S}_\alpha=\emptyset$ and $\cup_\alpha \mathbf{S}_\alpha=\mathbf{P}$. $M$ is the number of segments. The segment feature $\mathbf{f}_\alpha$ corresponding to $\mathbf{S}_\alpha$ is derived through an average pooling on $\mathbf{F}$:
\begin{equation}
    \mathbf{f}_\alpha=\mathop{\rm{avgpooling}}_{i\in \mathbf{S}_\alpha} (\mathbf{f}_i).
\end{equation}
For AGContrast, the set of positive pairs $\mathcal{P}=\{(i,\alpha)\ |\ i=1,2,\cdots,N\ \text{and}\ i\in \mathbf{S}_\alpha \}$, and the set of negative pairs $\mathcal{N}=\{(i,\beta)\ |\ i\notin \mathbf{S}_\beta\}$. Correspondingly, AGContrast can be defined as
\begin{equation}
    \mathcal{L}_{\rm{AG}}=-\sum_{(i,\alpha)\in\mathcal{P}}\log \frac{\exp (\mathbf{f}^1_i\cdot {\mathbf{f}^2_\alpha}^\top/\tau)}{\sum_{(i,\beta)\in \mathcal{N}}\exp (\mathbf{f}^1_i\cdot {\mathbf{f}^2_\beta}^\top/\tau)}.
\end{equation}

Enforcing asymmetric granularity contrastive learning between points and segments during pre-training confers several advantages: (1) \emph{Shape Content Perception.} Utilizing segment features to align with point features facilitates the assimilation of shape information during pre-training. This capability proves pivotal in tackling complex point cloud understanding tasks. (2) \emph{Avoidance of Close Similar Points as Negatives.} The selection of negative example samples for a given point $i$ excludes other points within the same segment $\mathbf{S}_\alpha$. This exclusionary criterion helps prevent the imposition of negative example loss on points that are already sufficiently similar. (3) \emph{Efficient Space Complexity.} Remarkably, AGContrast achieves point-level comparison learning with space complexity of only $O(MN)$, which is significantly reduced. This efficient utilization of resources is a noteworthy advantage of the approach.

\subsection{ChannelContrast}
The channel redundancy concern makes the network fall into a local optimum under unsupervised discriminative pretext task, which hinders the learning of inductive bias. To mitigate channel redundancy and foster robust inter-channel dependencies, we propose a channel-wise contrastive learning mechanism, which is illustrated in Fig.~\ref{fig:network} (b). Let $\mathbf{c}_i\in\mathbb{R}^{N\times 1}$ represent the $i$-th feature map of $\mathbf{F}$. ChannelContrast can be formally defined as follows:
\begin{equation}
    \mathcal{L}_{\rm{CC}}=-\sum_{(i,i)\in\mathcal{P}_c}\log \frac{\exp ({\mathbf{c}^1_i}^\top\cdot \mathbf{c}^2_i/\tau)}{\sum_{(i,j)\in \mathcal{N}_c}\exp (|{\mathbf{c}^1_i}^\top \cdot \mathbf{c}^2_j|/\tau)},
\label{eq:channelcontrast}
\end{equation}
where $\mathcal{P}_c=\{(i,i)\ |\ i=1,2,\cdots,C\}$ denotes the set of positive pairs and $\mathcal{N}=\{(i,j)\ |\ i\neq j\}$ denotes the set of negative pairs. Different from conventional contrastive loss, ChannelContrast considers the absolute value of the cosine distance from the query to the negative keys. Consequently, Eq.~\ref{eq:channelcontrast} promotes orthogonality between the feature map $\mathbf{c}^1_i$ and its corresponding negative feature map $\mathbf{c}^2_j$, which greatly mitigates channel redundancy.

Enforcing explicit contrastive learning within the channel dimensions offers several advantages: (1) \emph{Efficient Inductive Bias Learning.} The dynamic activation of feature maps across each dimension enhances the efficiency of inductive bias learning with $O(C^2)$ space complexity. (2) \emph{Fine-grained Supervision.} All points within the feature map actively contribute to the computation of the contrastive loss, ensuring that ChannelContrast maintains fine-grained supervision. (3) \emph{Reduction of Feature Dimension Redundancy.} ChannelContrast effectively mitigates feature dimension redundancy, paving the way for subsequent supervised training that makes efficient use of features.

\subsection{Loss Function}
The overall loss function for EPContrast is formed by combining the two components using a weighted sum operation, as follows:
\begin{equation}
\mathcal{L}_{\rm{EP}}=\mathcal{L}_{\rm{AG}}+\lambda\mathcal{L}_{\rm{CC}},
\label{eq:EPContrast}
\end{equation}
where $\lambda$ is the balance hyperparameter. With Eq.~\ref{eq:EPContrast}, EPContrast provides an efficient point-level contrastive learning for large-scale point cloud understanding.

\section{Experiments}

\subsection{Implementation}
\noindent \textbf{Datasets.} We evaluate EPContrast on two large-scale datasets, S3DIS~\cite{armeni20163d} and ScanNetV2~\cite{dai2017scannet}. S3DIS is a large-scene indoor 3D point cloud dataset, containing six instructional and office areas. The dataset has a total of nearly 700 million 3D points with color information, as well as semantic and instance labels assigned to each point. ScanNetV2 is an annotated rich point cloud dataset for 3D parsing of indoor scenes. It contains 1513 scans covering more than 700 unique indoor scenes, of which 1201 scans belong to the training set and the remaining 312 scans belong to the validation set.

\noindent
\textbf{Implementation Details.} We employ data augmentation techniques to create additional instances of the entire point cloud scene, which include scaling, rotating, and jittering of the point cloud scene. Furthermore, we employ unsupervised K-means clustering to derive segment-level results by grouping points based on their positional coordinates and color information. Specifically, we generate 2000 segments per scene through this clustering process. For a fair comparison, we set the number of sampled negative pairs to 2000 in other contrastive learning competitors as well. The temperature $\tau$ and the balance weight $\lambda$ are set to $1$ and $0.1$, respectively.

\begin{table*}[htbp]
\begin{center}
\caption{Comparisons on semantic segmentation, object detection, and instance segmentation downstream tasks. For fair comparisons, all methods are reproduced using the same settings on the same task. Scratch denotes training from scratch.}
\label{tab:comparison}
\begin{tabular}{lc|ccc|cc|cc}
\toprule[3\arrayrulewidth]
\multicolumn{2}{c|}{\begin{tabular}[c]{@{}c@{}}Task\\ (Backbone)\end{tabular}} & \multicolumn{3}{c|}{\begin{tabular}[c]{@{}c@{}}Semantic Segmentation\\ (PointNet++~\cite{qi2017pointnet++})\end{tabular}} & \multicolumn{2}{c|}{\begin{tabular}[c]{@{}c@{}}Object Detection\\ (VoteNet~\cite{qi2019vote})\end{tabular}} & \multicolumn{2}{c}{\begin{tabular}[c]{@{}c@{}}Instance Segmentation\\ (Sparse Res-U-Net~\cite{xie2020pointcontrast})\end{tabular}} \\ \hline
Method                                                   &Granularity &  mIoU                            & mACC                           & OA                             & mAP@0.5                                      & mAP@0.25                                     & mAP@0.5                                       & mAP@0.25                                       \\ \hline
Scratch & -                                                                        & 61.0                           & 86.8                          & 68.1                          & 35.4                                        & -                                           & 56.9                                         & -                                             \\ \hline
Info3d~\cite{sanghi2020info3d}                                                                    & Scene-level &  60.9                           & 86.8                          & 67.6                          & 35.7                                           & 57.7                                           & 56.5                                            & 70.3                                             \\
ToThePoint~\cite{li2023tothepoint}                                                                    & Scene-level & 61.0                           & 86.6                          & 67.7                          & 35.9                                           & 58.3                                           & 56.7                                            & 70.0                                             \\ \hline
SegContrast~\cite{nunes2022segcontrast}                                                                    & Segment-level &  61.4                           & 87.8                          & 68.1                          & 36.9                                           & 60.0                                           & 57.0                                            & 71.5                                             \\
CSC~\cite{hou2021exploring}                                                                           & Segment-level &  62.5                               & 87.4                              & 70.5                              & 38.8                                       & 60.8                                       & 57.5                                         &  \textbf{72.2}                                         \\ \hline
PointContrast~\cite{xie2020pointcontrast}                                                                  &Point-level &  61.9                           & 87.3                           & 68.7                          & 37.9                                       & 60.8                                        & 57.8                                         & 72.0                                          \\
MID~\cite{wang2021unsupervised}                                                                  &Point-level &  61.5                           & 87.1                           & 67.9                          & 37.1                                       & 59.9                                        & 57.4                                         & 71.3                                          \\
\rowcolor{green!15} EPContrast                                                                   &Point-level&  \textbf{62.8}                        & \textbf{87.9}                          & \textbf{70.9}                          & \textbf{39.8}                                        & \textbf{62.1}                                       & \textbf{58.6}                                             & 71.9                                         \\
\specialrule{0em}{1pt}{1pt}
\toprule[3\arrayrulewidth]
\end{tabular}
\end{center}
\end{table*}

\begin{table}[htbp]
\begin{center}
\caption{Ablation study for EPContrast on S3DIS semantic segmentation. ${\mathcal{L}_\text{PC}}^\star$ denotes the point-level contrastive loss without randomly sampling the negative pairs. OET denotes the One-Epoch Training.}
\label{tab:ablation}
\begin{tabular}{l|ccc}
\toprule[3\arrayrulewidth]
Method                            & mIoU (\%)     & OET (\%) & Memory \\ \hline
Scratch                           & 61.02    & 30.67 & -             \\
${\mathcal{L}_\text{PC}}^\star$                     & -    & - & 4395.2M             \\
$\mathcal{L}_\text{PC}$                     & 61.87    & 42.12 & 366.2M             \\
$\mathcal{L}_\text{PC}+\lambda\mathcal{L}_\text{CC}$     & 62.06    & 38.34       & 366.3M      \\
$\mathcal{L}_\text{AG}$                       & 62.75    & 50.66  & 368.2M            \\ 
\rowcolor{green!15}$\mathcal{L}_\text{AG}+\lambda\mathcal{L}_\text{CC}$                     & \textbf{62.81}    & \textbf{52.37} & 368.3M \\
\specialrule{0em}{1pt}{1pt}
\toprule[3\arrayrulewidth]
\end{tabular}
\end{center}
\end{table}

\noindent
\textbf{Downstream Tasks.} EPContrast undergoes comprehensive evaluation across three distinct downstream tasks: semantic segmentation, object detection, and instance segmentation. The key experimental details are thoughtfully documented in Tab.~\ref{tab:settings}. It is worth supplementing that the implementation of pointnet++~\cite{qi2017pointnet++} in semantic segmentation is followed by PointNeXt~\cite{qian2022pointnext}, the color information is not used in object detection following VoteNet~\cite{qi2019vote}, and the voxel size is 0.2cm for Sparse Res-U-Net~\cite{xie2020pointcontrast} in instance segmentation, in which the SparseConv is implemented by MinkowskiEngine~\cite{choy20194d}.

\subsection{Point Cloud Understanding}
We report performance comparisons with other contrastive learning methods on three downstream tasks. Comprehensive comparisons are illustrated in Tab.~\ref{tab:comparison}.

For semantic segmentation on S3DIS, EPContrast demonstrates notable improvements, with gains of 1.8\% mIoU, 1.1\% mACC, and 2.8\% OA when compared to learning from scratch. Compared with scene-level and segment-level contrastive learning, EPContrast exhibits superior point cloud comprehension capabilities. This enhanced performance is attributed to EPContrast's ability to establish fine-grained correspondences between points and segments. In the challenging domain of point cloud object detection, the advantages of EPContrast become even more pronounced. With a mIoU threshold of 0.5, EPContrast achieves a remarkable improvement of 4.4\% mAP compared to learning from scratch. Moreover, EPContrast shows excellent performance under the instance segmentation task with the more powerful Sparse Res-U-Net structure~\cite{xie2020pointcontrast}.

\subsection{Ablation Study}
Tab.~\ref{tab:ablation} shows the ablation studies about AGContrast and ChannelContrast on S3DIS semantic segmentation. Replacing $\mathcal{L}_{\rm PC}$ with AGContrast $\mathcal{L}_{\rm AG}$ brings 0.9\% mIoU performance improvement. The introduction of ChannelContrast $\mathcal{L}_{\rm CC}$ on $\mathcal{L}_{\rm PC}$ improves the semantic segmentation performance by 0.3\%. Furthermore, it's worth highlighting that EPContrast maintains a memory usage that is nearly equivalent to $\mathcal{L}_{\rm PC}$ after randomly sampling negative pairs. Without sampling negative pairs, $\mathcal{L}_{\rm PC}$ requires a huge amount of memory resources and therefore could not be implemented for pre-training in our experimental setup.

\subsection{One Epoch Training}

In addition, we perform an ablation study under One-Epoch Training, which means that the network learns the point cloud scene only once. As shown in Tab.~\ref{tab:ablation}, more than 21.7\% mIoU improvement compared to Scratch implies that EPContrast has significant application potential in few-shot tasks.

\subsection{Label Efficient Learning}
EPContrast enables label-efficient learning under sparse annotations~\cite{xu2020weakly}, as shown in Tab.~\ref{tab:label_effcient}. At extreme labeling rates of 0.01\% and 0.1\%, EPContrast gains 11.0\% and 2.1\% mIoU improvement over Scratch, respectively.  

\begin{table}[htbp]
\begin{center}
\caption{Label efficient study on S3DIS semantic segmentation.}
\label{tab:label_effcient}
\begin{tabular}{l|ccc}
\toprule[3\arrayrulewidth]
Method                            & mIoU (0.01\% label)    & mIoU (0.1\% label) \\ \hline
scratch                           & 43.59    & 58.30	              \\
Info3d                            & 44.61    &   58.86                  \\
SegContrast                       & 47.17    &   59.26                   \\
PointContrast                     & 51.17    & 60.22              \\
\rowcolor{green!15}EPContrast                      & \textbf{54.58}    & \textbf{60.36}  \\
\specialrule{0em}{1pt}{1pt}
\toprule[3\arrayrulewidth]
\end{tabular}
\end{center}
\end{table}
\section{Conclusion}
In this paper, we alleviate the tension between the reliance on pre-training for large point cloud scenarios and the excessive growth in spatial complexity for contrastive learning. By combining the proposed AGContrast and ChannelContrast, EPContrast maintains fine-grained supervised signals while effectively reducing the memory footprint and thus achieves better performance in various downstream tasks. In addition, EPContrast also outperforms other contrastive learning methods in one-round training and labeling efficient settings. We expect EPContrast to inspire more memory-efficient pre-training methods for large-scale point cloud understanding.

{
    \bibliographystyle{IEEEbib}
    \bibliography{icme2023template}
}

\end{document}